\documentclass[runningheads]{llncs}
\usepackage{amsmath}
\usepackage{graphicx}
\usepackage{multirow}
\usepackage{subfigure}
\usepackage{hyperref}
\usepackage[subtle,margins=normal,leading=normal]{savetrees} 
\begin{document}
%
\title{ConDistFL: Conditional Distillation for Federated Learning from Partially Annotated Data}
\authorrunning{P. Wang et al.}
\titlerunning{ConDistFL}
%
%
\author{Pochuan Wang \inst{1} \and
Chen Shen \inst{2} \and
Weichung Wang \inst{1}\and
Masahiro Oda \inst{2} \and
Chiou-Shann Fuh \inst{1} \and
Kensaku Mori \inst{2} \and
Holger R. Roth \inst{3} 
}
%
%
\institute{National Taiwan University, Taiwan \and
Nagoya University, Japan \and
NVIDIA Corporation, United States\\
\email{hroth@nvidia.com}
}
\maketitle              
%

\begin{abstract}

Developing a generalized segmentation model capable of simultaneously delineating multiple organs and diseases is highly desirable. Federated learning (FL) is a key technology enabling the collaborative development of a model without exchanging training data. However, the limited access to fully annotated training data poses a major challenge to training generalizable models.
We propose ``ConDistFL", a framework to solve this problem by combining FL with knowledge distillation. Local models can extract the knowledge of unlabeled organs and tumors from partially annotated data from the global model with an adequately designed conditional probability representation. We validate our framework on four distinct partially annotated abdominal CT datasets from the MSD and KiTS19 challenges. The experimental results show that the proposed framework significantly outperforms FedAvg and FedOpt baselines. Moreover, the performance on an external test dataset demonstrates superior generalizability compared to models trained on each dataset separately. Our ablation study suggests that ConDistFL can perform well without frequent aggregation, reducing the communication cost of FL. Our implementation will be available at \url{https://github.com/NVIDIA/NVFlare/tree/dev/research/condist-fl}.


\end{abstract}

\section{Introduction}

%
%
Accurately segmenting abdominal organs and malignancies from computed tomography (CT) scans is crucial for clinical applications such as computer-aided diagnosis and therapy planning.
While significant research has focused on segmenting individual organs~\cite{altini2022liver,gul2022deep} and multiple classes of organs without malignancies~\cite{cerrolaza2019Computational,fu2021review}, a generalized model capable of handling multiple organs and diseases simultaneously is desirable in real-world healthcare scenarios.
Traditional supervised learning methods, on the other hand, rely on the amount and quality of the training data. Regrettably, the cost of high-quality medical image data contributed to a paucity of training data. For many anatomies, only trained professionals can produce accurate annotations on medical images. On top of this, even experts often only have specialized knowledge for a specific task, making it challenging to annotate the organs and corresponding malignancies of various anatomies and imaging modalities. 

The lack of sufficient annotated datasets for multiple organs and tumors poses a significant challenge in developing generalized segmentation models. To address this issue, several studies have explored partially annotated datasets, where only a subset of targeted organs and malignancies are annotated in each image, to build generalized segmentation models~\cite{Shi2020-me,hongdong2022multi,fang2020multi,huang2020multi,tajbakhsh2020embracing}. However, sharing private medical datasets among institutions raises privacy and regulatory concerns. To overcome these challenges, federated learning (FL) was introduced~\cite{McMahan2017}. FL enables collaborative training of a shared (or ``global'') model across multiple institutions without centralizing the data in one location.

\begin{figure}[htb]
  \centering
  \includegraphics[width=\textwidth]{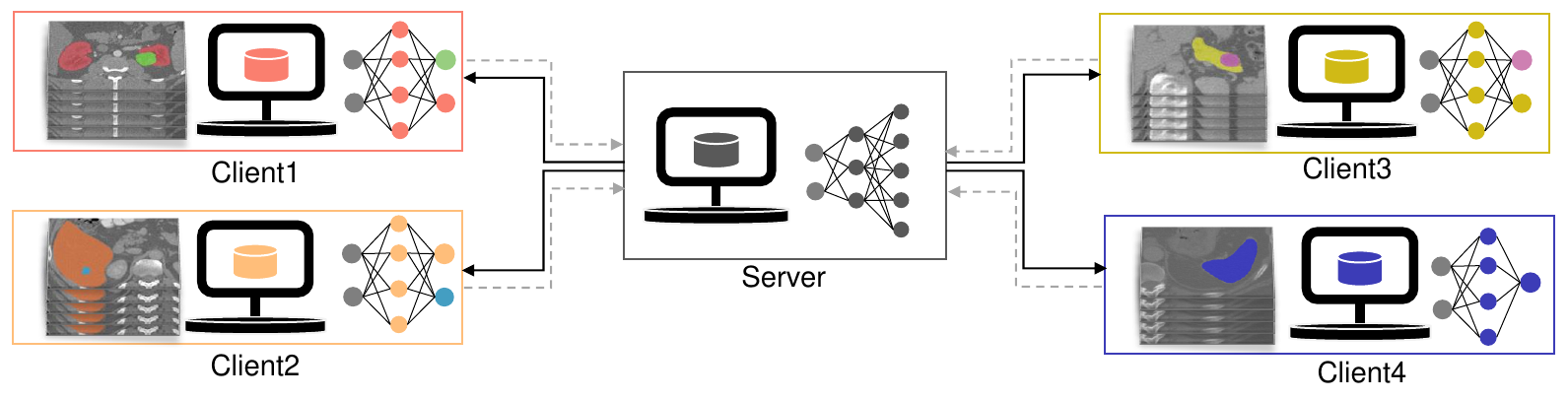}
  \caption{An illustration of the ConDistFL framework for multi-organ and tumor segmentation from partial labels. Each client has only a subset of the targeted organs and malignancies annotated in their local datasets.}
  \label{fig:condist_client}
\end{figure}

FL has emerged as a promising technology to enhance the efficiency of medical image segmentation \cite{sheller2020federated,xu2021federated,tajbakhsh2021guest}. In FL, each client trains a local model using its data and resources while only sending model updates to the server. The server then combines these updates into a global model using ``FedAvg''~\cite{McMahan2017}.
Recent studies have utilized FL to develop unified multi-organ segmentation models using partially annotated abdominal datasets~\cite{liuMultisiteOrganSegmentation2023,xu2022federated} as illustrated in Fig.~\ref{fig:condist_client}. However, these approaches often neglect lesion areas. Only a few studies attempt to generate to segment the various organs and their cancers simultaneously \cite{Zhang_2021_CVPR,shen2022joint}. The model aggregation in FL is a major hurdle because of the data heterogeneity problem brought on by data diversity~\cite{zhao2018federated}. Merging models from different sources with non-IID data can lead to performance degradation. This issue is further exacerbated when clients use data annotated for different tasks, introducing more domain shifts in the label space. Additionally, unbalanced dataset sizes among clients may affect the global model's performance on tasks with limited data.

In this work, we suggest a framework to tackle data heterogeneity in FL for multi-class organ and tumor segmentation from partially annotated abdominal CT images. The main contributions of this work are as follows:
\begin{enumerate}
    \item  Our proposed conditional distillation federated learning (ConDistFL) framework enables joint multi-task segmentation of abdominal organs and malignancies without additional fully annotated datasets.
    \item The proposed framework exhibits stability and performance with long local training steps and a limited number of aggregations, reducing data traffic and training time in real-world FL scenarios.
    \item We further validate our models on an unseen fully annotated public dataset AMOS22 \cite{ji2022amos}. The robustness of our approach is supported by both the qualitative and quantitative evaluation results.
\end{enumerate}

\section{Method}
%
ConDistFL extends the horizontal FL paradigm~\cite{yang2019federated} to handle partially annotated datasets distributed across clients. An illustration of our ConDistFL framework for multi-organ and tumor segmentation from partial labels is shown in Fig.~\ref{fig:condist_client}. In the client training of ConDistFL, we combine supervised learning on ground truth labels and knowledge distillation learning~\cite{Hinton2015-mu} using the global model's predictions. During supervised learning, we adopt the design of marginal loss~\cite{Shi2020-me} to avoid knowledge conflicts caused by missing labels. To improve the knowledge distillation in FL settings, we proposed a conditional distillation loss to maximize the agreement between the global model and local model predictions on unlabeled voxels.

\subsection{Conditional Distillation for Federated Learning}

\begin{figure}[htb]
  \centering
  \includegraphics[width=0.9\textwidth]{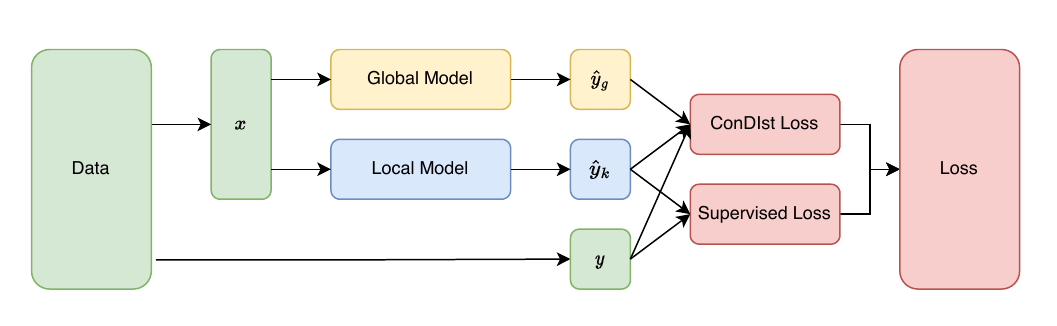}
  \caption{ConDistFL data flow diagram for client $k$; $x$ is a batch of image patches from the local dataset; $y$ is the corresponding label; $\hat{y}_{g}$ is the output of global model; and $\hat{y}_{k}$ is the output of the local model.}
  \label{fig:condist_data_flow}
\end{figure}

In ConDistFL, the client keeps the latest global model as the teacher model for knowledge distillation and uses the local model as the student model. Figure~\ref{fig:condist_data_flow} illustrates the training data flow in client $k$ and the relationship between the global, local, and loss functions.


\subsection{Supervised Loss}
\label{supervised-loss}

We adopt the design of marginal loss \cite{Shi2020-me} for the supervised loss $\mathcal{L}_{sup}$. Let $N$ be the total number of classes across all datasets, $F_k$ be a collection of foreground classes on client $k$, $B_k$ be the background class, and all unlabeled classes on client $k$, and $\hat{y}_{k, i}$ be the output logits of client $k$'s model for class $i$. By applying softmax normalization on the output logits $\hat{y}_{k, i}$ for each $i \in N$, we can derive the output probability $\hat{p}_{k, i}$ for each class $i$ as 

\begin{equation}
\label{eq:softmax}
  \hat{p}_{k,i} = \frac{e^{\hat{y}_{k,i}}}{\sum_{j=0}^N e^{\hat{y}_{k,j}}}\ \text{for}\ i=0,~1,~2,~\dots,~N-1.
\end{equation}
Similar to the marginal loss, all probabilities in the background $B_k$ are merged into one for a new non-foreground class. The probabilities remain the same as $\hat{p}_{k,i}$ for all $i \in F_k$. Then we apply Dice loss \cite{milletari2016v} with cross-entropy loss \cite{yi2004automated} (DiceCELoss) for supervised learning of the segmentation model. The final loss term $\mathcal{L}_{sup}$ is defined as

\begin{equation}
\label{eq:sup_loss}
  \mathcal{L}_{sup} = \textnormal{DiceCELoss}(\hat{p}'_{k}, y'),
\end{equation}
where the background merged probability is $\hat{p}'_k$, and the corresponding one-hot label is $y'$.

\subsection{ConDist Loss}

For the conditional distillation (ConDist) loss $\mathcal{L}_{ConDist}$, we normalize the output logits of both the global and the local model using softmax with temperature $\tau$. The normalized logits from the global model $\hat{p}^{\tau}_{g}$ and the $k$-th client's local model $\hat{p}^{\tau}_{k}$ is defined as
\begin{equation}
    \hat{p}^{\tau}_{k} = \textnormal{softmax}(\hat{y}_{k} / \tau)
    \quad \mathrm{and} \quad
    \hat{p}^{\tau}_{g} = \textnormal{softmax}(\hat{y}_{g} / \tau),
\end{equation}
where $\tau$ is set to $0.5$ to enhance the confidence of the model output.
\subsubsection{Foreground Merging and Background Grouping.}

Contrary to the supervised loss, we merge the probabilities for class $i$ for all $i \in F_{k}$ in $\hat{p}^{\tau}_{k}$ and $\hat{p}^{\tau}_{g}$. Then we define

\begin{equation}
\label{eq:foreground-merging}
    \hat{p}_{k, F_{k}} = \sum_{i \in F_{k}} \hat{p}^{\tau}_{k, i}
    \quad \mathrm{and} \quad
    \hat{p}_{g, F_{k}} = \sum_{i \in F_{k}} \hat{p}^{\tau}_{g, i},
\end{equation}
where the $\hat{p}^{\tau}_{k, i}$ and $\hat{p}^{\tau}_{g, i}$ are the probabilities for class $i$ in $\hat{p}^{\tau}_{k}$ and $\hat{p}^{\tau}_{g}$, respectively.
Moreover, we group the background classes in client $k$ by the organs in the background $B_{k}$. Let $M_{k}$ be the number of unlabeled organs in client $k$, and $\mathcal{O}_{k}=\{ G_{0}, G_{1}, G_{2}, \hdots G_{M_{k}} \}$. $G_{0}$ is the set containing the global background class. The probability for each background group can be calculated as

\begin{equation}
\label{eq:background-groups}
    \hat{p}_{k, G_{i}} = \sum_{j \in G_{i}} \hat{p}^{\tau}_{k, j}
    \quad \mathrm{and} \quad
    \hat{p}_{g, G_{i}} = \sum_{j \in G_{i}} \hat{p}^{\tau}_{g, j}
    \quad \mathrm{for} \quad
    i=0,~1,~\hdots,~M_{k},
\end{equation}
where $G_{i}$ is a set containing the class of the healthy part of unlabeled organ $i$ and all the classes of associated lesions to the organ $i$.
\subsubsection{Conditional Probability for Background Organs.}
We define conditional probabilities $\hat{p}_{k, \mathcal{O}_{k} | B_{k}}$ and $\hat{p}_{g, \mathcal{O}_{k} | B_{k}}$ as

\begin{align}
    \hat{p}_{k, \mathcal{O}_{k} | B_{k}} = \left(
        \frac{\hat{p}_{k, G_{0}}}{1-\hat{p}_{k, F_{k}}},
        \frac{\hat{p}_{k, G_{1}}}{1-\hat{p}_{k, F_{k}}},
        \hdots,
        \frac{\hat{p}_{k, G_{M_{k}}}}{1-\hat{p}_{k, F_{k}}}
    \right), \\
    \hat{p}_{g, \mathcal{O}_{k} | B_{k}} = \left(
        \frac{\hat{p}_{g, G_{0}}}{1-\hat{p}_{g, F_{k}}},
        \frac{\hat{p}_{g, G_{1}}}{1-\hat{p}_{g, F_{k}}},
        \hdots,
        \frac{\hat{p}_{g, G_{M_{k}}}}{1-\hat{p}_{g, F_{k}}}
    \right),
\end{align}
where $1 - \hat{p}_{k, F_{k}}$ and $1-\hat{p}_{g, F_{k}}$ are the total probabilities of all classes in $B_{k}$ with respect to $\hat{p}^{\tau}_{k}$ and $\hat{p}^{\tau}_{g}$. The conditional probability $\hat{p}_{k, \mathcal{O}_{k} | B_{k}}$ and $\hat{p}_{g, \mathcal{O}_{k} | B_{k}}$ are the probabilities given that the prediction is only in $B_{k}$.

\subsubsection{Foreground Filtering.}
To avoid learning from incorrect predictions and reduce the potential conflict with the supervised loss, we filter off undesired voxels with a mask operation $\mathcal{M}$, which removes the union of foreground area in ground truth label $y$ and all the area in $\hat{y}_{g}$ where the predictions are in $F_k$.

\subsubsection{Segmentation ConDist Loss}
Combining the conditional probability $\hat{p}_{k, \mathcal{O}_{k} | B_{k}}$, $\hat{p}_{g, \mathcal{O}_{k} | B_{k}}$, and the foreground filtering mask $\mathcal{M}$, we define the ConDist loss $\mathcal{L}_{ConDist}$ for segmentation task by applying soft Dice loss as

\begin{equation}
\label{eq:condist_loss}
    \mathcal{L}_{ConDist} = DiceLoss(
        \mathcal{M}(\hat{p}_{k, \mathcal{O}_{k} | B_{k}}),
        \mathcal{M}(\hat{p}_{g, \mathcal{O}_{k} | B_{k}})
).
\end{equation}

To handle meaningless global model predictions in the initial FL rounds, we incorporate an adjustable weight $w$ for the ConDist loss, gradually increasing it as the FL round number increments. The total loss $\mathcal{L}$ for ConDistFL is defined as

\begin{equation}
  \mathcal{L} = \mathcal{L}_{sup} + w * \mathcal{L}_{ConDist}.
\label{eq:loss_weight}
\end{equation}
In practice, we schedule the weight $w$ from $0.01$ to $1.0$ linearly.

\section{Experiments}

We conducted our experiments on the Medical Segmentation Decathlon (MSD) \cite{msd2} and the KiTS19 Challenge \cite{kits19} datasets. In the MSD dataset, we only used the liver, pancreas, and spleen subsets. Except for the spleen dataset, each above dataset includes annotations for the organs and tumors. We split the dataset into training, validation, and testing subsets by $60\%$, $20\%$, and $20\%$, respectively.
For the non-FL standalone training, each model only uses a single dataset. For FL training, we distributed the four datasets to four independent clients. 
In addition, we evaluated our models on the multi-modality Abdominal Multi-Organ Segmentation Challenge 2022 (AMOS22) dataset \cite{ji2022amos}, which consists of 300 CT volumes with 15 abdominal organ annotations. To accommodate the labeling format of AMOS22, where healthy organs and lesions are not distinguished, we merged the tumor predictions with associated organs from our model output before computing the metrics using ground truth labels.


The nnU-Net~\cite{Isensee2021-ur} data preprocessing pipeline was adopted with minor modifications. We first resampled the images to a median spacing of $[1.44,~1.44,~2.87]$ millimeters and clipped the intensity to the range $[-54,~258]$. Then we applied z-score normalization by assuming the mean intensity under ROI to be $100$ and its standard deviation to be $50$ since the complete multi-organ ROI is unavailable. We set the input patch size to $[224,~224,~64]$ and the training batch size to $4$.

Our neural network backbone was built using the 3D DynU-Net from MONAI~\cite{Jorge_Cardoso2022-wb}, an effective and flexible U-Net implementation.
Deep supervision was also enabled to speed up the training process and enhance model performance. The deep supervision loss is identical to the supervised loss with an extra exponential weight decay.
We trained our models using stochastic gradient descent (SGD) and cosine annealing schedule. The initial learning rate was set to $10^{-2}$ and decreased gradually to $10^{-7}$.

The loss function for the non-FL standalone baselines is Dice loss with cross-entropy. For the FL experiments, we evaluated FedAvg~\cite{McMahan2017}, FedOpt~\cite{asad2020fedopt}, FedProx~\cite{li2020federated}, and ConDistFL. To assess the effectiveness of the marginal loss in section \ref{supervised-loss}, we trained two sets of FedAvg models: one using the standard Dice loss and the other employing the marginal loss. The FedProx model was trained with the FedAvg aggregator and $\mu=0.01$. For FedOpt and ConDistFL, we utilized the Federated Optimization (FedOpt) aggregation method, with an additional SGD optimizer with momentum $m=0.6$ implemented on the server.

We employed a single NVIDIA V100 GPU for each standalone experiment and FL client. The FL experiments were implemented using NVIDIA FLARE~\cite{Roth2022-dk}.

\section{Results \& Discussion}
Our experiments encompass standalone baselines using a single dataset, an ablation study of standard Dice loss on FedAvg (FedAvg*), marginal loss on FedAvg, FedProx, and FedOpt, and the combined marginal loss and ConDist loss on ConDistFL. To establish a fair comparison between related works, we trained a ConDistFL (Union) model with the same learning targets as \cite{liuMultisiteOrganSegmentation2023} and \cite{xu2022federated}, i.e., only to segment the union of the organs and tumors. Additionally, we evaluated the proposed method on the unseen AMOS22 dataset to demonstrate its generalizability.

Table~\ref{tab:big-table} compares the average Dice score of each task between standalone baseline models and the best performance server models for FedAvg, FedProx, FedOpt, and ConDistFL.
All the models are trained for a total of $120,000$ steps to allow for a fair comparison. For the best FedAvg*, FedAvg, and ConDistFL models, we utilized $60$ FL aggregation rounds with $2000$ local steps per round. As for the FedProx and FedOpt best model, we conducted $120$ FL rounds and $1000$ local steps per round.

The results of FedAvg* and FedAvg demonstrate that the marginal loss effectively resolves conflicts between inconsistent labels and yields reasonable performance. ConDistFL stands out as the top-performing method among all experiments utilizing the marginal loss. FedAvg, FedProx, and FedOpt show similar performance overall, with FedAvg and FedOpt delivering acceptable results for most tasks, except for the pancreas and tumor. In contrast, FedProx performs well for the pancreas and tumor, but there is a notable drop in performance on other tasks. 
This suggests that although the FedProx loss can regularize the models for heterogeneous clients like the proposed ConDist loss, its task-agnostic nature harms the performance when training on multiple tasks with different partial labels.

\begin{table}[tb]
\centering
\caption{Comparison between non-FL results and each FL result. The average Dice score of each organ for the standalone model is computed separately from four distinct models. FedAvg* indicates the model trained with FedAvg and standard Dice loss.}
\begin{tabular}{l|cc|cc|cc|c|c}
\hline
  & Kidney & \multicolumn{1}{c|}{Tumor} & \multicolumn{1}{c}{Liver} & \multicolumn{1}{c|}{Tumor}
             & Pancreas & Tumor & Spleen & Average \\ \hline
  Standalone & 0.9563 & 0.8117
             & 0.9525 & 0.7071
             & 0.7974 & 0.5012
             & 0.9632 & 0.8102 \\ \hline\hline
  FedAvg*    & 0.7707 & 0.4894
             & 0.4937 & 0.3202
             & 0.5403 & 0.1396
             & 0.0000 & 0.3934 \\
  FedAvg     & 0.9419 & 0.6690
             & 0.9381 & 0.6500
             & 0.6933 & 0.2985
             & 0.9059 & 0.7281 \\
  FedProx    & 0.9247 & 0.6799
             & 0.8972 & 0.6244
             & 0.7419 & \textbf{0.4033}
             & 0.7060 & 0.7111 \\
  FedOpt     & 0.9473 & 0.7212
             & 0.9386 & 0.6087
             & 0.6734 & 0.2390
             & 0.9394 & 0.7239 \\
  ConDistFL  & \textbf{0.9477} & \textbf{0.7333}
             & \textbf{0.9446} & \textbf{0.6944}
             & \textbf{0.7478} & 0.3660
             & \textbf{0.9562} & \textbf{0.7700} \\
  \hline
\end{tabular}

\label{tab:big-table}
\end{table}

\begin{figure}[tb]
    \centering
    \includegraphics[width=0.9\textwidth]{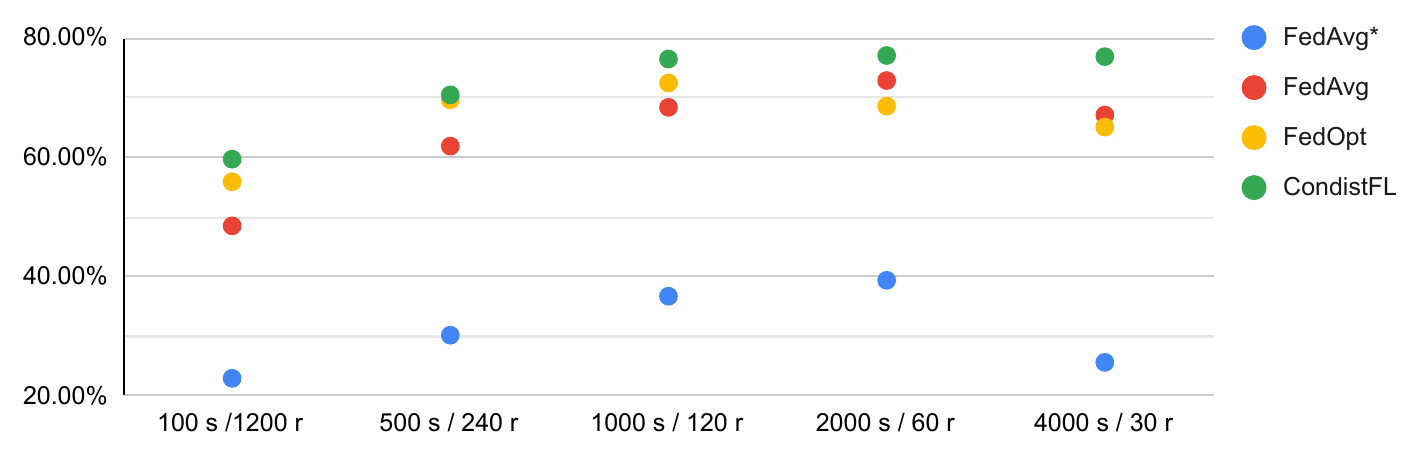}
    \caption{The ablation study results on the test set. The x-axis is the number of local training steps (s) and rounds numbers (r), while the y-axis is the average Dice score of all organs and tumors.}
    \label{fig:ablation}
\end{figure}

The ablation study in Fig.~\ref{fig:ablation} investigates the impact of the number of local training steps on the global model performance. Increasing the number of local training steps from $100$ to $1000$ for all tested methods improved performance. However, when more local training steps were used, both FedAvg and FedOpt encountered model divergence issues, with FedAvg experiencing a more significant performance drop than FedOpt. In contrast, ConDistFL consistently delivered better performance across different local step experiments. This can be attributed to ConDistFL providing a common task, preventing model divergence in FL, and the inherent complexity of tumor segmentation requiring more local training steps. By maintaining consistent representations of unknown classes, ConDistFL allows for the use of larger local step sizes to learn the tumor segmentation task effectively.

Table~\ref{tab:external-test} compares the average Dice scores of standalone baselines, ConDistFL, and ConDistFL (Union) on the unseen AMOS22 dataset. ConDistFL demonstrates significant generalizability improvements over the standalone baselines, while ConDistFL (Union) further enhances performance. 
This highlights the challenge of segmenting tumors and organs together compared to considering them a single class.

\begin{table}[tb]
\centering
\caption{External test results for AMOS22 dataset in average Dice scores.}
\begin{tabular}{l|ccccc}
\hline
                  & Kidney & Liver  & Pancreas & Spleen & Average \\ \hline
Standalone        & 0.5916 & 0.9419 & 0.5944 & 0.8388 & 0.7417 \\
FedAvg            & 0.5032 & 0.8718 & 0.4637 & 0.5768 & 0.6039 \\
FedProx           & 0.4698 & 0.6994 & 0.5185 & 0.7120 & 0.5999 \\
FedOpt            & 0.5171 & 0.6740 & 0.4113 & 0.6418 & 0.5611 \\
ConDistFL         & 0.7218 & 0.9191 & 0.6188 & 0.8556 & 0.7788 \\
ConDistFL (Union) & \textbf{0.8746} & \textbf{0.9471}
                  & \textbf{0.7401} & \textbf{0.9079} & \textbf{0.8674} \\
\hline
\end{tabular}
\label{tab:external-test}
\end{table}

Table~\ref{tab:related-works} compares ConDistFL (Union) and the reported performance of FL PSMOS and MENU-Net on the MSD test set. FL PSMOS utilizes a fully annotated dataset for model pre-training, while MENU-Net introduces a fifth client with a fully annotated dataset. The results demonstrate that ConDistFL achieves comparable performance without needing fully annotated data. Additionally, ConDistFL significantly reduces the number of aggregation rounds, leading to substantial savings in data traffic and synchronization overheads.

\begin{table}[tb]
\centering
\caption{Comparing the average Dice scores of ConDistFL to reported performance of related works. ``Rounds'' is the number of FL aggregation rounds.}
\begin{tabular}{l|c|ccccc}
\hline
       & Rounds
       & Kidney & Liver & Pancreas & Spleen & Average \\ \hline
FL PSMOS \cite{liuMultisiteOrganSegmentation2023}
       & 2,000
       & \textbf{0.966}  & 0.938  & 0.788  & \textbf{0.965}  & 0.9143 \\
MENU-Net \cite{xu2022federated}
       & 400
       & 0.9594 & 0.9407 & 0.8005 & 0.9465 & 0.9118 \\
ConDistFL (Union)
       & 120
       & 0.9657 & \textbf{0.9619} & \textbf{0.8210} & 0.9626 & \textbf{0.9278} \\
\hline
\end{tabular}
\label{tab:related-works}
\end{table}


\begin{figure}[tb]
\centering
\subfigure{\includegraphics[clip,width=0.6\textwidth]{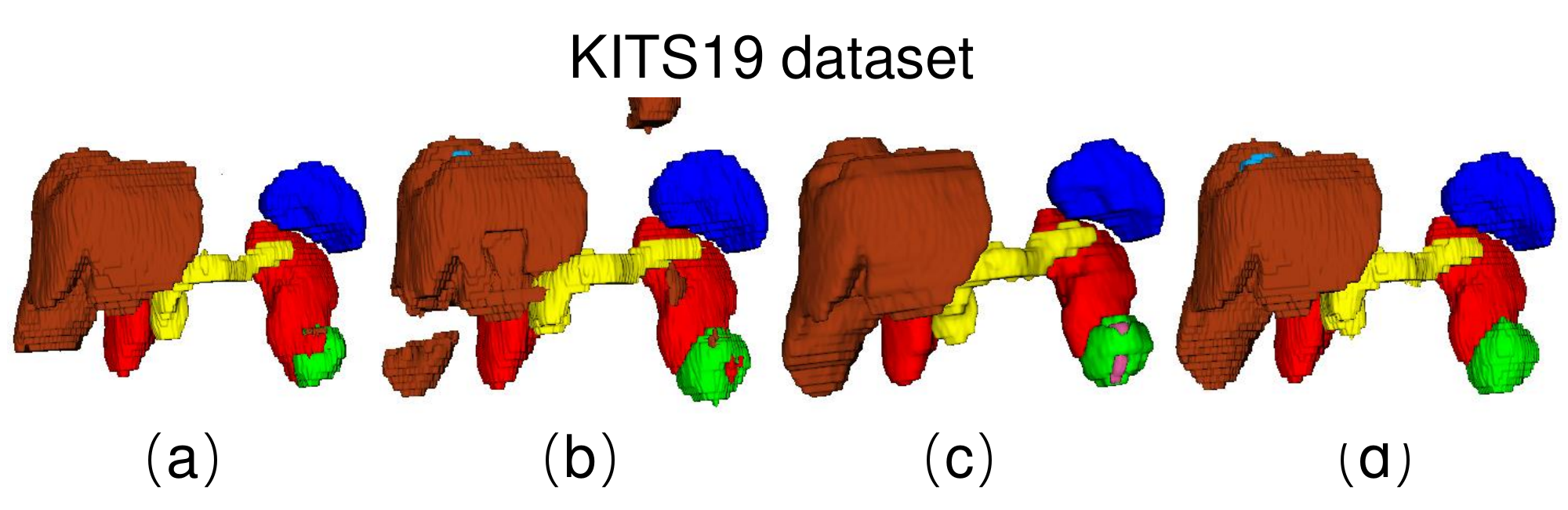}}
\subfigure{\includegraphics[clip,width=0.3\textwidth]{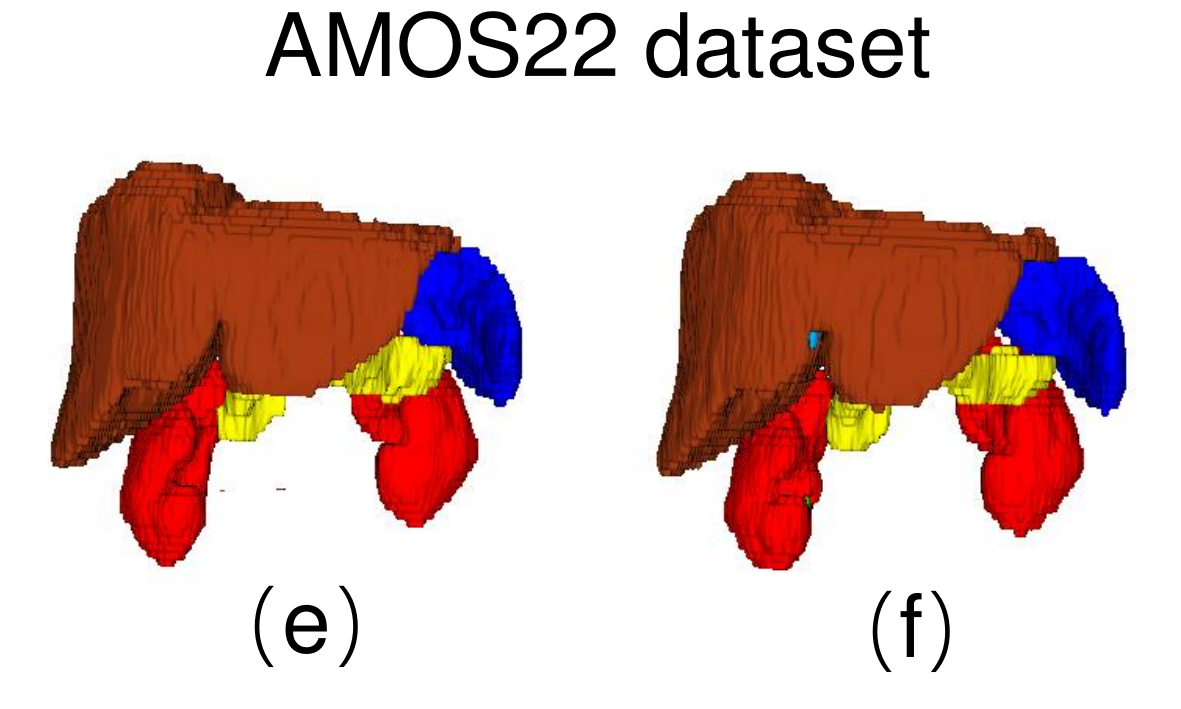}}
\includegraphics[width=0.9\textwidth]{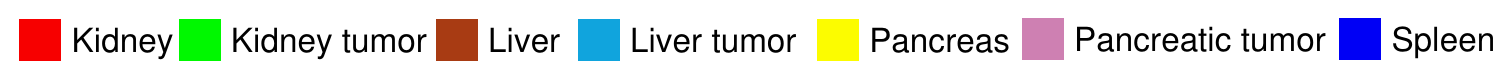}
\caption{3D renderings of segmentation on the best performed FL server model using (a) FedAvg, (b) FedOpt, (c) FedProx, (d) ConDistFL on KITS19 data, and (e) ground truth and (f) the external segmentation using ConDistFL on AMOS22 data.
}
\label{fig:results}
\end{figure}
Fig.~\ref{fig:results} showcases 3D visualizations of our proposed ConDistFL, demonstrating effective and simultaneous segmentation of multiple organs and tumors without ensembling. Compared to FedAvg and FedOpt, ConDistFL achieves smoother and more continuous segmentations. The comparison with the ground truth of AMOS22 validates the generalizability of our FL framework.
\section{Conclusion}
This work offers a promising FL approach for generalized segmentation models from partially annotated abdominal organs and tumors, reducing annotation costs and speeding up model development. Moreover, the proposed method requires less frequent aggregation, making it suitable for real-world FL scenarios with limited communication bandwidth.

%
%

\bibliographystyle{splncs04}
\bibliography{references}
\end{document}